\documentclass[11pt]{article}

\usepackage[preprint]{acl}

\usepackage{times}
\usepackage{latexsym}
\usepackage[T1]{fontenc}
\usepackage[utf8]{inputenc}
\usepackage{microtype}
\usepackage{inconsolata}
\usepackage{graphicx}
\usepackage{booktabs}
\usepackage{multirow}
\usepackage{xcolor}
\usepackage{amsmath}
\usepackage{marvosym}
\usepackage{CJKutf8}

\newcommand{\zh}[1]{\begin{CJK}{UTF8}{gbsn}#1\end{CJK}}

\title{Slogans or Stance? \\
       A Label-Light Diagnostic for Entrepreneurial-Discourse \\
       Measurement on Chinese SOE Speeches}

\author{
Ting Gong, Shangquan Sun\textsuperscript{\Letter}\\
Tsinghua University\\
\texttt{gongting@umich.edu}
}

\begin{document}
\maketitle

\begin{abstract}
Dictionary methods, topic models, and embedding-similarity scorers
are widely used in CSS and management research to measure constructs
such as ``entrepreneurial spirit'' in corporate speeches.  We
contribute a \emph{label-light measurement diagnostic} for such
instruments rather than a new extraction model.  On a corpus of 80
speeches by leaders of centrally administered Chinese state-owned
enterprises, we exploit a natural experiment of 24 same-company
different-speaker pairs and 5 same-company same-speaker pairs to
test whether a method's per-document indices vary with leader
identity holding firm constant.  LDA fails (Cohen $d=0.20$, 95\% CI
$[-0.72, 1.20]$); a dictionary scorer reaches $d=0.81$ and a Chinese
sentence encoder $d=0.65$ on doc-vector distances of order $10^{-3}$.
A zero-shot 9B open-weight LLM (Qwen3.5:9b) raises paired-contrast
$d$ to $1.09$ (exact permutation $p_1=0.034$).  We downgrade three
claims accordingly: gold $F_1$ measures consistency with the LLM's
own prompt rule rather than external construct recovery; doc-level
style residualisation cuts the LLM's $d$ to $0.43$ ($p_1=0.22$), so
roughly half of the effect is consistent with leader idiolect; and a
confidence-weighted calibration trades $\Delta$ for variance with an
auto-mined slogan lexicon near-inert in ablation.  We release the
2{,}190-segment scored corpus, the 170-paragraph pilot, the slogan
lexicon, two-family LLM scores, and the evaluation harness.
\end{abstract}

\section{Introduction}
\label{sec:intro}

Empirical work on entrepreneurship and corporate governance relies
on textual indices derived from corporate disclosures and leadership
addresses.  Such indices serve as inputs to studies of firm
performance, innovation, and political affiliation, and they often
rest on one of three method families: keyword dictionaries
\cite{loughran2011liability,baker2016measuring,huang2020measuring},
latent topic models \cite{blei2003latent}, and pre-trained word or
sentence embeddings \cite{mikolov2013distributed,devlin2019bert}.  All three measure topical \emph{coverage}: how
much a document discusses innovation, risk, or sustainability.  None
was designed to separate the speaker's \emph{substantive} stance on a
construct from the speaker's \emph{symbolic-rhetorical} performance
about it.

This separation matters in discourse genres where symbolic performance
is part of the institutional contract.  Speeches by leaders of
centrally administered Chinese state-owned enterprises (SOEs) are a
clear example.  An SOE chairperson's address is expected to include
canonical political invocations --- for example the policy slogan
\emph{``cultivate world-class enterprises with global competitiveness''}
or the doctrinal formulation \emph{``state-owned enterprises are an
important material and political foundation of socialism with Chinese
characteristics''} (see Table~\ref{tab:slogan} for further entries) ---
alongside operational reporting such as a specific divestiture, a
named joint venture, or a numerically anchored R\&D programme.  The
two coexist in the same speech, often in the same paragraph.  A
dictionary-based ``innovation index'' computed over such a speech is
largely driven by the former, because slogans contain most of the
keywords that a hand-curated dictionary recognises and are
interchangeable across speeches.

We therefore ask: \emph{do standard entrepreneurial-discourse extraction
methods, as widely deployed on Chinese corporate corpora, measure
leader-level stance, or do they measure recurrent political symbolism?}

To answer this without commissioning a new annotation round we exploit
a property of our corpus: 29 of 51 companies appear in two interview
waves.  Twenty-four of those wave pairs involve a \emph{change of
speaker} at the same company; five involve the \emph{same speaker}
reappearing.  If a method captures leader-level stance, its per-document
vectors should differ more across leader-change pairs than across
same-leader pairs.  If a method captures the company's industry topic
or shared political ritual, the two distributions should be
indistinguishable.

Our contributions are:

\begin{enumerate}
\item A \emph{leader-change paired evaluation} for stance extraction in
performative corporate discourse that requires no in-domain
annotation (§\ref{sec:eval}).

\item An audit of four method families on 80 central-SOE speeches that
shows that dictionary, LDA, and BGE-encoder methods fail the
evaluation: their paired-contrast effect either crosses zero in its
95\% bootstrap CI or is computed on doc-vector distances of order
$10^{-3}$ (§\ref{sec:results}).  A complementary \emph{paraphrase-
robustness} experiment confirms that the dictionary baseline
\emph{increases} its score when substantive content is rewritten in
slogan style (§\ref{sec:para}).

\item A \emph{confidence-weighted calibration} on a zero-shot 9B
open-weight LLM (Qwen3.5:9b) that trades two paired-contrast metrics:
it raises absolute $\Delta$ by 27\% but reduces Cohen's $d$ from
$1.09$ to $0.83$; ablation localises the lift to LLM self-confidence,
not to the auto-mined slogan lexicon.  Qualitative ordering is
preserved under a within-family cross-LLM check (Qwen3.5:27b,
§\ref{sec:crossllm}).

\item Release of the 2{,}190-segment scored corpus, the 170-paragraph
pilot gold set, the 53-entry auto-mined slogan lexicon, and the full
evaluation harness.
\end{enumerate}

\section{Related Work}
\label{sec:related}

\paragraph{Entrepreneurial orientation and entrepreneurial leadership.}
The management and entrepreneurship literature has long treated
entrepreneurial orientation (EO) as a structured construct of
innovativeness, proactiveness, risk-taking, autonomy, and competitive
aggressiveness \cite{miller1983correlates,covin1989strategic,
lumpkin1996clarifying}, and subsequent reviews emphasise both its
firm-level centrality and the instability of its operationalisation
across contexts \cite{rauch2009entrepreneurial,wales2016entrepreneurial}.
Parallel work on corporate entrepreneurship and entrepreneurial
leadership argues that such constructs are enacted through strategic
posture and leader discourse rather than through isolated keywords
\cite{dess2005role,kuratko1999corporate,harrison2016entrepreneurial,
bagheri2020entrepreneurial}.  Our paper inherits this measurement
target and asks whether standard text-as-data pipelines recover it
from SOE speeches.

\paragraph{Dictionary-based discourse measurement.}
The Loughran--McDonald financial sentiment dictionary
\cite{loughran2011liability}, Tetlock's media-sentiment work
\cite{tetlock2007giving}, and the Baker--Bloom--Davis economic-policy
uncertainty index \cite{baker2016measuring}, with its Chinese-language
extension \cite{huang2020measuring}, are emblematic of a research
programme that operationalises latent corporate-discourse constructs via
hand-curated word lists.  Within the Chinese-corporate literature, the
same template has been applied to ``entrepreneurial spirit'', innovation
orientation, and political-mission framing, typically as document-level
proportions of seed-word hits.  The implicit assumption is that the
distribution of construct-related vocabulary in a document is informative
about the document's underlying stance.  Our paired-contrast evaluation
makes this assumption testable: if the assumption holds, dictionary
scores must vary more across leaders than within them.  Our findings
(§\ref{sec:results}) indicate that, for SOE leadership speeches at least,
the assumption is empirically weak.

\paragraph{Topic models, hierarchical classification, and embedding similarity.}
Latent Dirichlet Allocation \cite{blei2003latent} is ubiquitous in CSS
analyses of long political and corporate text \cite{grimmer2013text,
roberts2014structural}.  Embedding-based topic models \cite{angelov2020top2vec}
generalise the idea to a continuous space; hierarchical multi-label
text classification \cite{chalkidis2020empirical,shen2021taxoclass,
xu2021hierarchical,falis2021cophe} is a closer match to our L1/L2
taxonomy but requires labelled training data, which is unavailable
here.  Modern Chinese sentence encoders such as BGE
\cite{xiao2023bge}, building on the Sentence-BERT line
\cite{reimers2019sentence}, achieve strong zero-shot retrieval
performance and are a common default for unsupervised dimension
scoring in CSS.  We
benchmark all three families and analyse why each underperforms on
our paired-contrast task: LDA recovers industry topics rather than
stance dimensions, and a domain-monolithic sentence encoder collapses
per-document distances on a homogeneous corpus
(§\ref{sec:results}, §\ref{sec:discussion}).

\paragraph{Stance, framing, and performative discourse.}
A long line of NLP work distinguishes stance from topic, mostly for
short-form social-media data \cite{mohammad2016semeval,allaway2020zero}.
Framing analysis in political communication
\cite{card2015media,field2018framing} treats discourse as a choice among
competing emphases rather than a position on a single axis, and several
recent papers extend it to corporate communication
\cite{ziems2024can}.  A separate strand of work in political theory and
linguistics treats institutional speech as \emph{performative}: an
utterance whose primary function is not to assert but to perform an
institutional act \cite{austin1962how,searle1969speech}.  SOE
leadership addresses are performative in this sense: the political
invocations they contain serve role enactment rather than information
transfer.  Our slogan/substance separation operationalises this
distinction quantitatively for a corporate-discourse setting in which
performative and substantive content coexist within the same paragraph.
This move also connects to entrepreneurship research that treats
entrepreneurial communication as rhetoric, discourse, and
meaning-making rather than as transparent disclosure
\cite{holt2010sensemaking,roundy2018themes,riedy2022discursive,
salmivaara2020rhetoric,caliskan2022entrepreneurialism,
steyaert2005narrative}.  Those studies motivate our central concern:
what appears as ``entrepreneurial'' language may partly reflect a
discursive template or institutional genre rather than the leader's
substantive strategic stance.

\paragraph{Construct validity in NLP measurement.}
\citet{jacobs2021measurement} argue that NLP-based measurement
instruments routinely conflate operational measurement with theoretical
construct, and call for measurement-validity diagnostics analogous to
those standard in psychometrics.  Our leader-change paired evaluation
contributes one such diagnostic for performative corporate discourse:
because the design holds the firm constant while varying the leader,
methods that score the firm's industry rather than the leader's stance
are unmasked without requiring construct-level labels.  Corporate-NLP
benchmarks for financial NLI, risk text, and ESG disclosure
\cite{mathur2022docfin,magomere2025finnli,tang2025finmteb,
he2025esgenius,padhi2024value} mostly treat the corpus as declarative
rather than performative.

\paragraph{LLMs as annotators and as measurement instruments.}
Frontier LLMs have been shown to match or exceed crowd-worker
agreement on stance and framing tasks \cite{gilardi2023chatgpt}, and
a growing literature uses them as scalable replacements for human
annotators or as judges and probes in CSS work
\cite{ziems2024can,heseltine2024llms,koval2024learning,
liscio2022cross,chuang2025judging}.  The question that
follows is whether downstream methods built on top of LLM scores
contribute beyond re-packaging the LLM.  Our ablation
(§\ref{sec:ablation}) addresses this directly: the slogan-aware
calibration term that contributes the most absolute lift is the
multiplier on the LLM's own substantive-confidence output, not the
externally mined slogan lexicon.  We interpret this as evidence that,
on this corpus, much of the value added by post-hoc calibration is
auditable re-weighting of confidence rather than orthogonal signal
recovery.

\section{Data}
\label{sec:data}

\paragraph{Corpus.}
We use 80 publicly released speeches delivered between 2018 and 2021
by leaders of centrally administered Chinese state-owned enterprises
and by officers of the State-owned Assets Supervision and
Administration Commission (SASAC).  After normalising firm names,
the 80 documents cover 51 unique organisations; 29 of these appear in both
interview waves.  Document-level statistics are summarised in
Table~\ref{tab:corpus}.

\begin{table}[t]
\centering
\small
\begin{tabular}{lr}
\toprule
\textbf{Statistic} & \textbf{Value} \\
\midrule
Documents                & 80 \\
Unique companies         & 51 \\
Wave-A only / Wave-B only & 22 / 22 \\
Both waves               & 29 \\
\quad same speaker (longitudinal) & 5 \\
\quad different speaker (leader change) & 24 \\
Segments (after splitting) & 2{,}190 \\
Median chars / document  & 10{,}741 \\
Median segments / document & 27 \\
\bottomrule
\end{tabular}
\vspace{-3mm}\caption{Corpus summary.}
\label{tab:corpus}
\end{table}

\paragraph{Segment construction.}
Each speech is split into paragraphs by blank lines.  Any paragraph
exceeding 600 characters is recursively split, first at enumeration
markers (e.g.\ \emph{first}, \emph{second}) and contrastive
connectives (e.g.\ \emph{however}, \emph{moreover}), and, if necessary,
at sentence boundaries.  Fragments shorter than 10 characters (likely
headings) are dropped.  The resulting 2{,}190 segments have a median
length of 525 characters (P90 = 589).

\paragraph{Gold pilot.}
A single annotator coded all paragraphs of five documents under a
written protocol.  Documents were stratified-sampled across leader
archetypes (one regulator, one operator-style chairperson, one
technology-driven firm, one CSR-themed firm, one capital-management
firm) to span the expected range of stance density.  The L1 label is
one of \textsc{slogan}, \textsc{substantive}, or \textsc{irrelevant},
with the operational rule \emph{``if the paragraph could be transplanted
verbatim into a different SOE speech and still read naturally, label it
\textsc{slogan}.''}  \textsc{Substantive} paragraphs additionally
received a dimension label (cf.\ §\ref{sec:method}).  The pilot yields
170 paragraph labels (74 \textsc{slogan}, 93 \textsc{substantive}, 3
\textsc{irrelevant}).  No method is trained on the pilot; it is a
small held-out test set for gold $F_1$ (§\ref{sec:eval}).  The
operational rule given to the annotator matches the rule passed to
the LLM in Appendix~\ref{app:prompt}, so gold $F_1$ is a rule-consistency
metric rather than a construct-validity check.

\paragraph{Pilot--segment alignment.}
The auto-segmenter is slightly more granular than the hand coding.
We align each annotated paragraph to a unique auto-segment by (i)
strict character-prefix match after whitespace removal, (ii) substring
fallback, and (iii) character-4-gram Jaccard fallback; each segment is
claimed by at most one paragraph.  Of the 170 paragraphs, 86 align with
match score $\geq 0.7$; gold $F_1$ is computed on the 83 of these whose
L1 is \textsc{slogan} or \textsc{substantive}.  Lower-confidence matches
are excluded; §\ref{sec:limitations} returns to this.

\section{Methods}
\label{sec:method}

\subsection{The five dimensions}

We follow the operationalisation of entrepreneurial spirit in
\citet{gong2024unveiling}: five coarse-grained dimensions, each
associated with 25 Chinese seed words.  We refer to them as
\textit{Innovation}, \textit{Competition--Cooperation},
\textit{Organisation--Market}, \textit{Social Responsibility}, and
\textit{National Mission}.

This is a localised adaptation of the entrepreneurial orientation
(EO) construct \cite{miller1983correlates,covin1989strategic,
lumpkin1996clarifying}: \textit{Innovation} and
\textit{Competition--Cooperation} correspond to innovativeness and
competitive aggressiveness; \textit{Organisation--Market} broadly
covers proactiveness and autonomy via organisational/market-mechanism
reform; \textit{Social Responsibility} and \textit{National Mission}
are SOE-specific additions without direct EO counterparts.
Risk-taking is not scored separately because SOE addresses rarely
articulate risk preferences --- a known operationalisation gap
\cite{wales2016entrepreneurial}.

For non-dictionary methods we additionally provide a one-sentence
natural-language description per dimension that explicitly contrasts
substantive realisation with slogan realisation
(e.g.\ \textit{Innovation}: ``concrete R\&D investment, patent
results, key-technology breakthroughs, new product or new business
launches; does not include the abstract slogan \emph{adhere to
innovation-driven development}'').  These descriptions serve as
anchor texts for the sentence-encoder baseline and as context for the
LLM prompt.

\subsection{Baselines}

\paragraph{Dictionary (\textsc{Dict}).}
Per segment, we tokenise with the jieba Chinese segmenter and
report, for each dimension, the fraction of tokens that match the
dimension's seed-word set.

\paragraph{Latent Dirichlet allocation (\textsc{Lda}).}
We train a gensim LDA with $K=5$ topics on the 80 documents and map
each topic to a dimension by greedy seed-word argmax (ties broken by
raw score).  Per-segment scores are projected topic probabilities
under this mapping.  This vanilla configuration is intentionally a
weak baseline; seeded LDA, larger $K$ with merging, or STM with
leader covariates may do better and are left to future work.

\paragraph{Sentence-encoder cosine (\textsc{Bge}).}
We embed each segment and each dimension description with the
\texttt{BAAI/bge-small-zh-v1.5} encoder \cite{xiao2023bge} and use the
cosine similarity as the dimension score.

\subsection{LLM extractor (\textsc{Llm})}

We prompt Qwen3.5:9b \cite{qwen2024technical} running locally under
Ollama in its native generate endpoint with thinking disabled.  The
prompt asks for a JSON object with four fields, which we will refer
to throughout by their short names: a stance-type label $L_1 \in
\{\textsc{slogan}, \textsc{substantive}, \textsc{irrelevant}\}$ with
the same operational rule given to the human annotator; a substantive
confidence $c_{\text{sub}} \in [0,1]$ giving the probability that the
segment is substantive; per-dimension stance scores
$s_{\text{raw}}(d) \in [0,1]$, each \emph{conditional on the segment
being substantive}; and a slogan density
$\rho_{\text{llm}} \in [0,1]$, the model's estimate of the character
fraction occupied by political symbolism.  The compositional structure
of the stance scores and the substantive confidence is deliberate:
it lets us compute either the raw \textsc{Llm} reading (stance
conditional on being substantive) or a substance-weighted reading
(stance discounted by how likely the segment is to be substantive at
all), which is the basis of the calibration below.

\subsection{Confidence-weighted calibration (\textsc{Calibrated})}

We weight the LLM's per-dimension stance score by its self-reported
substantive confidence and by two slogan-density penalties; the
ablation in §\ref{sec:ablation} attributes essentially all of the
paired-contrast lift to the confidence multiplier and shows the
slogan-density terms to be near-inert (we retain them for
interpretability).  The product is a single-source weak-supervision
aggregator \cite{pruthi2020weakly,mekala2020meta}.  In parallel we
compute a corpus-derived \emph{n-gram slogan density} per segment: a
weighted character-overlap with a lexicon mined from cross-document
n-gram repetition.  The
lexicon (53 entries) comprises every jieba 5-gram that appears in at
least 15\% of the 80 documents; each entry is weighted by
$\log(\mathrm{df}+1)$, where $\mathrm{df}$ is its document frequency.
The mining procedure recovers, without supervision, the recurring
political set phrases catalogued in Table~\ref{tab:slogan}.

\begin{table}[t]
\centering
\small
\begin{tabular}{p{0.45\columnwidth} p{0.40\columnwidth} r}
\toprule
\textbf{Original phrase} & \textbf{Literal English gloss} & \textbf{df} \\
\midrule
\zh{培育具有全球竞争力的世界一流企业} &
cultivate world-class enterprises with global competitiveness & 44 \\
\zh{党的十九大报告} &
report of the 19th Party Congress & 36 \\
\zh{习近平新时代中国特色社会主义思想} &
Xi Jinping Thought on Socialism with Chinese Characteristics for a New Era & 29 \\
\zh{实现中华民族伟大复兴} &
realise the great rejuvenation of the Chinese nation & 26 \\
\zh{国有企业是中国特色社会主义的重要物质基础和政治基础} &
state-owned enterprises are an important material and political foundation of socialism with Chinese characteristics & 20 \\
\zh{党执政兴国的重要支柱和依靠力量} &
an important pillar and reliable force for the Party's rule and the nation's revitalisation & 16 \\
\bottomrule
\end{tabular}
\vspace{-3mm}\caption{Six representative entries from the 53-entry auto-mined slogan
lexicon, ranked by document frequency ($\mathrm{df}$, out of 80
documents).  English glosses are literal translations provided for
non-Chinese readers and are not used by the model.}
\label{tab:slogan}
\end{table}

The calibrated per-dimension score is:
\begin{equation}
s_{\text{cal}}(d) = s_{\text{raw}}(d)
\,\cdot\, c_{\text{sub}}
\,\cdot\, m_{\text{llm}}
\,\cdot\, m_{\text{ng}}
\end{equation}
with $m_{\text{llm}} = \max(0, 1 - \lambda_{\text{llm}}\,\rho_{\text{llm}})$
and $m_{\text{ng}} = \max(0, 1 - \lambda_{\text{ng}}\,\rho_{\text{ng}})$,
where $s_{\text{raw}}(d)$ is the LLM's stance score for dimension
$d$, $c_{\text{sub}}$ is the LLM's substantive confidence,
$\rho_{\text{llm}}$ is the LLM-reported slogan density, and
$\rho_{\text{ng}}$ is the n-gram-derived slogan density.  We report two configurations.  The first sets
$\lambda_{\text{llm}} = \lambda_{\text{ng}} = 0$ (i.e.\ confidence-only
calibration), which the $\lambda$ grid search in §\ref{sec:results}
identifies as the optimum on the paired-contrast Cohen $d$ landscape.
The second sets $\lambda_{\text{llm}} = 1.0$, $\lambda_{\text{ng}} = 2.0$
(full calibration), which retains the slogan-density multipliers for
interpretability at a small cost in $d$.  We refer to the full
calibration as \textsc{Calibrated} throughout.

For all five methods, the per-document dimension vector is the
arithmetic mean of segment scores per dimension over that document's
segments; no further normalisation is applied before evaluation.

\section{Evaluation}
\label{sec:eval}

\subsection{Channel 1: leader-change paired contrast}

For each method we form two paired distance distributions over per-document
dimension vectors:

\begin{description}
\item[$D_{\text{lc}}$] = $\{\cos\text{-dist}(v_{A}, v_{B})\}$ for each of the
24 same-company, different-speaker A/B pairs.
\item[$D_{\text{sl}}$] = $\{\cos\text{-dist}(v_{A}, v_{B})\}$ for each of the
5 same-company, same-speaker A/B pairs.
\end{description}

We report the means, the absolute difference $\Delta = \bar D_{\text{lc}} -
\bar D_{\text{sl}}$, Cohen's $d$ with pooled within-group standard
deviation \cite{cohen2013statistical}, and a Hedges' $g$ small-sample
correction.  Two complementary uncertainty channels are reported:
bootstrap CIs with 2{,}000 resamples
\cite{tibshirani1993introduction} (subject to known small-$n$
limitations at $n_{\text{sl}}=5$) and exact permutation $p$-values
under the null that pair labels are exchangeable, enumerating all
$\binom{29}{5}=118{,}755$ partitions.

A method that captures leader-specific stance should produce
$\bar D_{\text{lc}} \gg \bar D_{\text{sl}}$.  A method that captures the
company's industry or topic vocabulary should produce
$\bar D_{\text{lc}} \approx \bar D_{\text{sl}}$, because both pairs hold the
company constant.

\subsection{Channel 2: gold-pilot binary $F_1$}

On the 83-paragraph gold subset, each method predicts an L1 label
in $\{\textsc{slogan}, \textsc{substantive}\}$ and we report $F_1$
on the substantive class.  For \textsc{Llm} and \textsc{Calibrated}
we use the model's predicted L1 directly.  \textsc{Dict}, \textsc{Lda},
and \textsc{Bge} have no native verdict, so we threshold their
maximum-dimension score at the median over the 2{,}190 segments; the
ordering of methods is robust to threshold choice in our experiments.
This channel should be read alongside paired-contrast
(§\ref{sec:results}) and paraphrase-robustness (§\ref{sec:para}):
the gold L1 rule matches the LLM prompt rule, so $F_1$ here is
rule-consistency, not construct-validity, and the median threshold
fixes baseline positive rate at 50\% by construction against a
$\approx 55$\% gold base rate.

External behavioural validity (per-firm per-year correlation with patent
counts, R\&D intensity, or ESG scores) is the natural third evaluation
channel for this construct.  We did not collect those signals for the
present corpus; we discuss what completing that step would add in
§\ref{sec:limitations}.

\section{Results}
\label{sec:results}

\subsection{Main results}

\begin{figure}[t]
\centering
\includegraphics[width=\columnwidth]{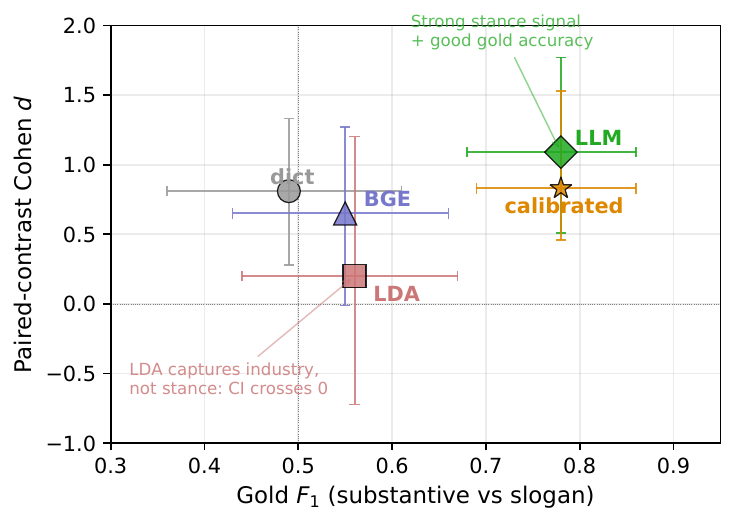}
\vspace{-3mm}\caption{Method comparison on the two evaluation channels.  Markers are
point estimates; error bars are 2{,}000-resample bootstrap 95\% CIs.
\textsc{Lda}'s paired-contrast Cohen $d$ ($0.20$) has a CI that
includes zero; \textsc{Dict}, \textsc{Lda}, and \textsc{Bge} cluster
near $F_1 \approx 0.5$ on the gold channel; \textsc{Llm} and
\textsc{Calibrated} lead on both axes.  \textsc{Bge}'s $d$ of $0.65$
is computed on doc-vector distances of order $10^{-3}$; see
§\ref{sec:results}.}
\label{fig:main}
\end{figure}

Table~\ref{tab:main} and Figure~\ref{fig:main} report the headline
numbers.  We summarise four observations.

\begin{table*}[t]
\centering
\small
\begin{tabular}{lrrcccc}
\toprule
\textbf{Method} & $\Delta$ & $d$ ($g$) & 95\% CI on $d$ & perm $p_1$ & ROC-AUC & PR-AUC \\
\midrule
\textsc{Dict}        & 0.098 & 0.81 (0.78) & [0.28, 1.33] & 0.082 & 0.41 & 0.54 \\
\textsc{Lda}         & 0.065 & \textbf{0.20 (0.19)} & [$-$0.72, 1.20] & 0.353 & 0.45 & 0.52 \\
\textsc{Bge}         & 0.001 & 0.65 (0.64) & [$-$0.01, 1.27] & 0.132 & 0.37 & 0.52 \\
\textsc{Llm}         & 0.105 & \textbf{1.09 (1.06)} & [0.51, 1.77] & \textbf{0.034} & 0.65 & 0.63 \\
\textsc{Calibrated}  & \textbf{0.130} & 0.83 (0.81) & [0.46, 1.53] & \textbf{0.042} & \textbf{0.72} & \textbf{0.74} \\
\bottomrule
\end{tabular}
\vspace{-3mm}\caption{Main results.  $\bar D_{\text{lc}}$ and
$\bar D_{\text{sl}}$ are mean cosine distances over the 24
leader-change and 5 same-leader pairs (full $\bar D$ values:
\textsc{Dict} 0.147/0.049, \textsc{Lda} 0.213/0.148, \textsc{Bge}
0.001/0.001, \textsc{Llm} 0.149/0.044, \textsc{Calibrated} 0.179/0.049);
$\Delta = \bar D_{\text{lc}} - \bar D_{\text{sl}}$.  Cohen's $d$ uses
pooled SD; Hedges' $g$ (in parentheses) applies a small-sample
correction; 95\% CIs are 2{,}000-resample bootstrap.  $p_1$ is the
exact one-sided permutation $p$-value enumerating all
$\binom{29}{5}=118{,}755$ partitions under the null of exchangeable
pair labels.  ROC-AUC and PR-AUC are threshold-free measures on the
83-paragraph gold subset (substantive vs slogan).}
\label{tab:main}
\end{table*}

Three signals corroborate.  The exact permutation test gives
one-sided $p < 0.05$ for \textsc{Llm} ($0.034$) and \textsc{Calibrated}
($0.042$); the three baselines fail this one-sided threshold.
Threshold-free PR-AUC peaks at \textsc{Calibrated} ($0.74$); \textsc{Dict}
and \textsc{Bge} ROC-AUC sit below $0.5$, indicating
anti-correlation with the gold substantive label.

\paragraph{\textsc{Lda}'s paired-contrast effect is not
distinguishable from zero.}
The bootstrap 95\% CI on \textsc{Lda}'s Cohen $d$ is $[-0.72, 1.20]$
and includes zero (permutation $p_1=0.35$): under the leader-change
paradigm we cannot reject
the null that the per-document dimension vectors of a same-leader pair
are drawn from the same distribution as those of a leader-change pair.
Inspection of the trained topics is consistent with this finding: the
five topics correspond to industry clusters (oil and petrochemicals;
electricity and grid; automotive and transport; construction and
infrastructure; aerospace and military) rather than to the
entrepreneurial-spirit dimensions.  Projecting industry topics through
a seed-word mapping yields a per-document vector that varies with the
firm but not with the leader.

\paragraph{\textsc{Bge}'s paired-contrast effect is on doc-vector
distances of order $10^{-3}$.}
The \textsc{Bge} per-document vectors lie in a narrow cone of the
embedding space: $\bar D_{\text{lc}} = 0.0012$ and $\bar D_{\text{sl}}
= 0.0005$.  Although the resulting Cohen $d = 0.65$ would be regarded
as a medium effect on its own, the gap on which it is computed is two
orders of magnitude smaller than that of the other methods, leaving
no operating margin for downstream use.  We attribute this collapse to
the homogeneity of SOE speech vocabulary, against which a
domain-agnostic sentence encoder cannot resolve leader-level
variation.

\paragraph{\textsc{Llm} leads on both channels.}
Zero-shot Qwen3.5:9b reaches $F_1 = 0.78$ on the gold subset (CI
$[0.68, 0.86]$, non-overlapping with any baseline CI) and Cohen $d =
1.09$ on the paired contrast (CI $[0.51, 1.77]$).  Both effects are
large.  A natural interpretation is that the model's prompt-level
operationalisation of stance is already most of what is needed; the
ablation in §\ref{sec:ablation} sharpens this reading.

\paragraph{Calibration enlarges absolute paired-contrast difference but
not Cohen $d$.}
Slogan-aware calibration raises the absolute paired-contrast difference
$\Delta$ from $0.105$ to $0.130$ ($+27\%$); the same-leader distance
$\bar D_{\text{sl}}$ rises by only $0.005$ while the leader-change
distance $\bar D_{\text{lc}}$ rises by $0.030$, so the gap grows.
However, the within-group standard deviation also rises proportionally,
so Cohen $d$ drops from $1.09$ to $0.83$.  We report both metrics
explicitly: Cohen $d$ rewards methods whose paired distances are
internally stable, while absolute $\Delta$ rewards methods that
discriminate at usable magnitudes on the corpus.  For downstream uses
that compare per-document vectors at face value (e.g.\ leader
trajectories or firm-year aggregates), $\Delta$ is the more relevant
quantity.

\subsection{Per-dimension breakdown}
\label{sec:perdim}

\begin{figure}[t]
\centering
\includegraphics[width=\columnwidth]{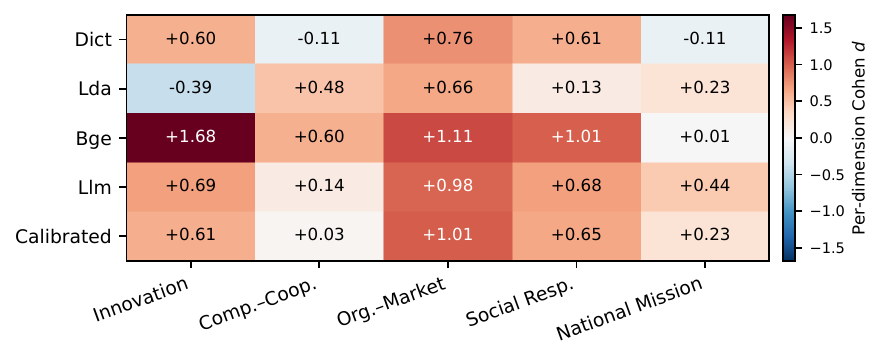}
\vspace{-3mm}\caption{Per-dimension Cohen $d$ for the leader-change vs same-leader
paired contrast, by method.  Larger and bluer cells indicate stronger
discrimination on that dimension; numeric values are the per-dimension
$d$.  \textsc{Llm} is uniformly positive across dimensions;
\textsc{Lda} is negative on \emph{Innovation} (leader-change pairs are
more similar than same-leader pairs in LDA's mapped innovation
distribution).  \textsc{Bge} has high Cohen $d$ but on the doc-vector
distances reported in Table~\ref{tab:main} the underlying gaps remain
$\sim 0.02$.}
\label{fig:perdim}
\end{figure}

Figure~\ref{fig:perdim} decomposes Table~\ref{tab:main}'s aggregate
$d$ into per-dimension contributions.  Three patterns are visible.
First, \emph{Organisation--Market} is the dimension on which every
method discriminates best ($d \geq 0.66$ across all methods), a
reflection of the fact that operational decisions (mixed-ownership
reform, listings, executive incentives) are the most concretely
realised stance content in our corpus.  Second, \textsc{Lda}'s
projection onto \emph{Innovation} is negatively correlated with the
leader contrast ($d = -0.39$): paragraphs about technical innovation
fall under the industry topics LDA discovers (aerospace, automotive,
etc.), so leader changes within a firm are barely visible.  Third,
\emph{National Mission} is the hardest dimension for every method;
\textsc{Llm} (0.44) leads, but the absolute differences are smaller
than for the other dimensions because national-mission discourse is
itself heavily formulaic.

\subsection{L2 sub-type analysis}
\label{sec:l2}

The LLM additionally classifies every \textsc{substantive} segment
into one of three sub-types: \textit{firm-action} (the leader's own
firm-specific decisions), \textit{policy-history} (recitation of
national policy and institutional history), or \textit{system-aggregate}
(industry- or system-level statistics, e.g.\ ``central SOEs cut
1.6\,Mt of steel capacity'').  Across the 1{,}419 \textsc{substantive}
segments, the distribution is 57\% \textit{firm-action}, 29\%
\textit{system-aggregate}, and 14\% \textit{policy-history}.  Per-doc
breakdowns are consistent with stereotype: an operator-style
chairperson's speech is dominated by \textit{firm-action} (e.g.\ 81\%
for the China National Building Materials chairperson in our pilot),
while a regulator's address contains a balanced mix.

Aggregating the L2 fractions to per-document vectors and re-running
the paired contrast gives a finer-grained answer to where leader
identity is visible: $\bar D_{\text{lc}} - \bar D_{\text{sl}} =
+0.144$ on the \textit{firm-action} fraction, but only $+0.060$ on
\textit{system-aggregate} and $+0.051$ on \textit{policy-history}.
The leader-change signal is concentrated in firm-action content, not
in policy or aggregate recitation --- a finding that is invisible if
one collapses all \textsc{substantive} content into a single channel.

\subsection{Component ablation}
\label{sec:ablation}

\begin{table}[t]
\centering
\small
\begin{tabular}{lcc}
\toprule
\textbf{Variant} & $\Delta$ & Cohen $d$ \\
\midrule
raw LLM (no calibration) & 0.105 & \textbf{1.09} \\
$\times \, c_{\text{sub}}$                          & 0.133 & 0.86 \\
$\times \,(1 - \rho_{\text{llm}})$                  & 0.101 & 1.03 \\
$\times \,(1 - 2\rho_{\text{ng}})$                  & 0.105 & 1.09 \\
$\times \,(1 - \rho_{\text{llm}})(1 - 2\rho_{\text{ng}})$ & 0.101 & 1.03 \\
all three (default \textsc{Calibrated})             & \textbf{0.130} & 0.83 \\
\bottomrule
\end{tabular}
\vspace{-3mm}\caption{Ablation of calibration components on paired-contrast.}
\label{tab:ablation}
\end{table}

The substantive-confidence multiplier $c_{\text{sub}}$ is the only
term that materially shifts $\Delta$; the two slogan-density
multipliers are near-inert in isolation and slightly over-correct
when combined with $c_{\text{sub}}$.

\subsection{Cross-model robustness}
\label{sec:crossllm}

We re-scored two 85-segment stratified subsamples under Qwen3.5:27b
(within-family) and DeepSeek-r1:8b (cross-family).  Within-family
agreement is high: Cohen's $\kappa$ \cite{cohen1960coefficient} is
$0.75$ on L1 and $0.51$ on L2; mean per-dim Pearson $r = 0.88$.
Cross-family agreement is substantial: $\kappa_{L1}=0.70$, mean
per-dim $r = 0.76$, substantive-confidence $r = 0.84$ (L2 degrades
to $\kappa = 0.30$).  These mitigate but do not eliminate
pretraining-overlap concerns (§\ref{sec:limitations}); full numbers
Appendix~\ref{app:crossllm}.

\subsection{Paraphrase robustness}
\label{sec:para}

We test how each method responds to \emph{surface-form change}.
For 50 segments Qwen3.5:9b labelled \textsc{substantive} with
confidence $\geq 0.6$, the same LLM rewrites each in slogan style
while preserving the claim; we then re-score.  Mean retention ratio
(rewrite/original) on the maximum-dimension score is $1.55$ for
\textsc{Dict}, $0.75$ for raw \textsc{Llm}, and $\mathbf{0.69}$ for
\textsc{Calibrated} (Appendix~\ref{app:para}).  \textsc{Dict}
\emph{rises} 33\% because rewrites are more keyword-dense;
\textsc{Calibrated} drops most because slogan density rises
($0.16 \!\to\! 0.40$) and substantive confidence falls
($0.95 \!\to\! 0.74$), propagating through the product.

\subsection{Sensitivity to small-$n$ SL and to leader style}
\label{sec:sensitivity}

Two sensitivity checks test how much of the LLM advantage survives
worst-case framing.  \textbf{Leave-one-SL-out}: \textsc{Llm}'s Cohen
$d$ stays in $[0.98, 1.25]$ (perm $p_1 \leq 0.071$) and
\textsc{Calibrated}'s in $[0.76, 0.94]$, while \textsc{Lda} swings
between $0.08$ and $0.93$ --- confirming the small-$n$ SL group is
the dominant source of LDA's wide CI.  \textbf{Style residualisation}:
regressing each method's per-document score on five doc-level
stylometric features (sentence-length mean/SD, numeric density,
long-run density, type-token ratio) cuts \textsc{Llm}'s Cohen $d$
from $1.09$ to $0.43$ ($p_1=0.22$) and \textsc{Calibrated}'s from
$0.83$ to $0.39$; absolute $\Delta$ stays positive ($0.22$ and
$0.19$).  Method ordering is preserved but no method is significant
post-residualisation, so roughly half of the LLM's measured
leader-change effect is consistent with leader style rather than
stance.  Full numbers in Appendix~\ref{app:sensitivity}.

\subsection{Where the methods diverge}
\label{sec:case}

A representative paragraph pair (Appendix~\ref{app:case})
illustrates the failure: a SASAC-chairperson political invocation
receives a higher dictionary \emph{Innovation} score than a CNBM
operational narrative receives for \emph{Organisation--Market}.  This
is systematic: across the gold pilot, \textsc{slogan} paragraphs
have mean dictionary max-score $0.071$ vs $0.054$ for
\textsc{substantive} ones --- dictionary scoring is anti-correlated
with the gold L1 label, the failure the calibration corrects.

A two-dimensional grid over $\lambda_{\text{llm}} \in \{0, 0.5, 1.0,
1.5, 2.0\}$ and $\lambda_{\text{ng}} \in \{0, 1, 2, 3\}$ produces
Cohen $d$ in the narrow range $[0.83, 0.86]$ (full grid in
Appendix~\ref{app:grid}).  The maximum is attained at
$\lambda_{\text{llm}} = 0$, consistent with the ablation finding that
the slogan-density multipliers contribute little once $c_{\text{sub}}$
is active.

\section{Discussion}
\label{sec:discussion}

\paragraph{Topic models and encoders recover industry, not stance.}
LDA's discovered topics align with industrial sectors rather than
with entrepreneurial-spirit dimensions; post-hoc re-labelling
inherits the original distributional semantics, not the new label.
\textsc{Bge}'s per-document vectors lie in a narrow cone of the
embedding space because every SOE speech shares the same corporate
and political vocabulary; dimension-aware anchor texts only mildly
mitigate this.

\paragraph{LLM advantage, with caveats.}
A 9B LLM picks up signal the three baselines miss
\cite{gilardi2023chatgpt}, but three caveats apply: gold L1 matches
the LLM prompt rule (§\ref{sec:limitations}); Qwen3.5's pretraining
likely overlaps the corpus; and roughly half the effect is consistent
with leader style (§\ref{sec:sensitivity}).  Calibration trades
$+27\%$ on $\Delta$ for $-0.26$ on Cohen $d$.  The diagnostic
transfers to corpora with performative-register discourse and
adequate speaker-rotation coverage (parliamentary speeches, earnings
calls).

\section{Conclusion}
\label{sec:conclusion}

On 80 Chinese SOE speeches, dictionary, LDA, and BGE produce weak
leader-specific signal; a zero-shot 9B LLM gives a larger contrast
($p_1=0.034$; cross-family $\kappa=0.70$), though half the effect is
consistent with leader style.  We frame the contribution as a
measurement-validity diagnostic and release the full evaluation
harness.

\clearpage
\section*{Limitations}
\label{sec:limitations}

\paragraph{Corpus scope.}
80 speeches across 51 firms is small.  All speeches are from central
SOEs in the 2018--2021 window; we do not claim our findings transfer
to provincial SOEs, private firms, or non-Chinese corporate discourse.
In particular our claim that BGE collapses is corpus-specific: on
more topically diverse corpora a domain-agnostic Chinese sentence
encoder might not collapse and the picture could improve.

\paragraph{Single-annotator gold pilot; no human-human IAA.}
Gold L1 verdicts come from one annotator on five documents; we do
not report inter-annotator agreement, and the coding rule was
written in tandem with the LLM prompt (Appendix~\ref{app:prompt}),
so gold $F_1$ measures rule consistency, not construct recovery.
The reported $F_1$ values are a ranking signal rather than precise
point estimates; a noise ceiling on this channel requires a second
independent annotator and would tighten the CIs.  Of 170 coded
paragraphs, 86 align to a unique auto-segment with match score
$\geq 0.7$; the rest are excluded from $F_1$.

\paragraph{Discriminative vs construct validity.}
Our paired contrast establishes that the LLM's per-document vectors
vary with leader identity holding firm constant; it does \emph{not}
establish that what varies corresponds to the theoretical EO
construct or to externally observable strategic posture.  A pure
stylometric baseline (eight features: mean paragraph length,
first-person rate, hedging, certainty, recurring-cliché frequency,
2-gram lexical diversity) yields paired contrast $d = 0.73$,
permutation $p_1 = 0.099$ on the same 24/5 pair structure --- below
\textsc{Llm}'s $d = 1.09$ and $p_1 = 0.034$, but large enough that
style is a non-trivial partial confound, not fully ruled out.
Throughout we use ``leader-attributable strategic stance'' for the
measured quantity and reserve ``entrepreneurial orientation'' for
the theoretical construct; bridging the two requires an external
behavioural channel (next paragraph).

\paragraph{Cross-LLM coverage and residual pretraining concerns.}
We checked robustness against Qwen3.5:27b (within-family) and
DeepSeek-r1:8b (cross-family).  Cross-family agreement is
substantially weaker than within-family ($\kappa_{L1}=0.69$ vs
$0.75$; mean per-dimension $r=0.76$ vs $0.88$; L2 $\kappa=0.30$ vs
$0.51$), and both models still share a heavy Chinese-pretraining
exposure plausibly including our corpus.  Replication with a
non-Chinese-native or older-cutoff model would tighten the
contamination argument further.  We also observed one anomaly --- a
single speech with all 28 segments labelled \textsc{slogan} ---
consistent with model-specific bias against the military-industry
register.

\paragraph{No external behavioural validity channel.}
The natural third channel for this construct is correlation between
per-firm per-year aggregate dimension scores and external behavioural
indicators (patent counts and R\&D intensity for \textit{Innovation};
ESG scores for \textit{Social Responsibility}).  Completing this
channel would require aligning the 51 firms to public IP-office and
financial-disclosure databases over the 2018--2021 window.  The
present paper claims only discriminative validity (a method's scores
distinguish leaders) and not construct validity (the scores
correspond to externally observable firm behaviour).

\paragraph{Sample sizes and group imbalance.}
The paired contrast rests on $n_{\text{lc}}=24$ leader-change pairs
versus $n_{\text{sl}}=5$ same-leader pairs.  The two groups are also
imbalanced on covariates: the SL group contains the SASAC regulator
(20\%; LC contains 0\%) and only two industry tags vs the LC group's
sixteen, which compresses the SL distribution toward a formulaic
register and almost certainly inflates the LC-vs-SL contrast.  The
exact permutation test partially addresses this by labelling under
exchangeability, but the underlying imbalance is intrinsic to the
natural experiment in this corpus and cannot be substantially
corrected without a larger SL pool.

\paragraph{Cohen $d$ versus absolute $\Delta$.}
Cohen $d$ is variance-controlled and conservative; absolute $\Delta$
is not.  We report both, but the paraphrase-robustness experiment
(§\ref{sec:para}) provides additional evidence that complements both
metrics: it identifies the method whose scores actually respond to
the substantive-vs-symbolic distinction at the level of an individual
paragraph rewrite.

\section*{Ethics statement}

The corpus consists of publicly released speeches and lectures by
senior officers of state-owned enterprises and a state regulator,
all of whom spoke in their official capacities.  We do not analyse
private communications or attribute material to any natural person
beyond their public professional role.

\paragraph{Terminology.}
``Slogan'' is a technical label for the performative-political
register following the operational rule given in §\ref{sec:data};
it is neither pejorative nor a normative judgement on Chinese
political discourse.  Equivalent terms in the literature include
\emph{performative-political register} and \emph{symbolic
invocation}; both refer to the same phenomenon and could replace our
label without loss.  Both performative and substantive content serve
legitimate communicative functions and our analysis does not endorse
a preference for either.

\paragraph{Name handling in released artefacts.}
Speaker names appear in our case-study (Appendix~\ref{app:case}) and
in segment metadata because the speeches are public and attributable
to public officials; we provide a name-redacted variant of the
corpus alongside the original for researchers who prefer it.  We
discourage repurposing the released scores for individual-leader
political-loyalty scoring or other ranking applications outside the
benchmark's intended measurement use.

The released benchmark may be of use to researchers in computational
social science studying performative corporate discourse, and to
management-science researchers seeking to validate or replace
existing dictionary-based measurement instruments.  We see no
foreseeable risk of harmful application beyond the routine risks of
any benchmark whose labels reflect a specific researcher's coding
protocol.

\section*{Reproducibility}

\paragraph{Resources released.}
Source code, the 2{,}190-segment scored corpus, the 170-paragraph
hand-coded pilot, the 53-entry auto-mined slogan lexicon, the
LLM-produced scores under both models, and all evaluation and
visualisation scripts are released at the URL given on the
camera-ready version.

\paragraph{Hardware and runtime.}
All experiments were run on a 48\,GB Apple M4 Pro workstation.  The
LLM extractor uses the Qwen3.5:9b open-weight model (Q4\_K\_M
quantisation) served by the Ollama runtime via its native generate
endpoint with thinking disabled, temperature $0$, and
\texttt{num\_predict}\,$=320$.  The full 2{,}190-segment scoring run
completes in approximately 30 minutes of wall-clock time; the
cross-LLM subsample under Qwen3.5:27b runs in approximately
3 minutes per segment.  No fine-tuning is performed; no hyperparameter
is tuned on the gold pilot.

\paragraph{Determinism and seeds.}
Bootstrap CIs use 2{,}000 resamples with a fixed Python random seed
(42).  The LLM is invoked at temperature 0 and is deterministic
modulo Ollama-internal scheduling; we re-ran the full corpus once and
verified identical L1 verdicts on a 100-segment spot-check.

\paragraph{Baseline implementations.}
The sentence-encoder baseline uses
\texttt{BAAI/bge-small-zh-v1.5} (Apache-2.0 licence; $\sim 10^8$
parameters); LDA uses the gensim implementation with $K=5$ topics,
10 passes, and 200 iterations; dictionary scoring uses the jieba
Chinese segmenter and the 25-keyword-per-dimension seed list of
\citet{gong2024unveiling}.  Dictionary scoring, LDA inference, the
n-gram lexicon mining, and the calibration step all run in seconds on
CPU and require no GPU.

\paragraph{Data licence.}
All speeches in the corpus were released by SASAC or the speakers'
own firms as part of the public ``SOE Open Lecture (100 Lectures)''
initiative and were collected from publicly available sources.
Speakers delivered the speeches in their official capacities.  We
redistribute only material that the original sources have made
public, with attribution.

\bibliography{custom,citaion_}

\appendix

\section{LLM extraction prompt (original Chinese)}
\label{app:prompt}

The prompt below is sent verbatim to Qwen3.5:9b for each segment.
\texttt{\{dim\_descriptions\}} is replaced by the five
short-form dimension descriptions defined in §\ref{sec:method};
\texttt{\{segment\_text\}} is replaced by the segment text.  The model
is invoked through the Ollama native \texttt{/api/generate} endpoint
with \texttt{think: false}, \texttt{temperature: 0}, and
\texttt{num\_predict: 320}.  We provide a literal English gloss
beneath each Chinese line for readers unfamiliar with the language.

\begin{quote}\small
\zh{分析下面这段中国国企领导讲话片段。仅输出 JSON。}\\
\textit{(Analyse the following segment of a Chinese SOE leadership speech.
Output JSON only.)}\\[3pt]
\zh{五个维度：} \texttt{\{dim\_descriptions\}}\\
\textit{(Five dimensions: [substituted at run time])}\\[3pt]
\zh{判定 l1：}\\
\textit{(Decide L1:)}\\
\zh{- slogan：抽象口号 / 政治表态 / 领导人引文 / 四字格 / 无具体主体-动词-数字-取舍。判别：可原封不动搬到任意国企讲话} $\to$ \zh{slogan}\\
\textit{(\textsc{slogan}: abstract slogans / political statements /
leader quotations / four-character formulas / no specific
subject-verb-number-tradeoff. Heuristic: if the paragraph could be
transplanted verbatim into any other SOE speech, label \textsc{slogan}.)}\\
\zh{- substantive：具体动词+客体 / 资源配置数字 / 可观测后果 / 对比取舍 / 时地特异性，满足 ≥2 项}\\
\textit{(\textsc{substantive}: at least two of \{specific verb+object,
resource allocation in numbers, observable outcome, contrast or tradeoff,
specificity of time/place/object\}.)}\\
\zh{- irrelevant：寒暄、过渡、与 5 维度均无关}\\
\textit{(\textsc{irrelevant}: greetings, transitions, not related to any
of the five dimensions.)}\\[3pt]
\zh{l2（仅 substantive 填）：firm\_action（本企业 leader 决策）/ policy\_history（政策制度叙事）/ system\_aggregate（系统聚合统计）}\\
\textit{(L2, only when L1 = \textsc{substantive}: firm\_action /
policy\_history / system\_aggregate; defined in §\ref{sec:l2}.)}\\[3pt]
\zh{输出严格 JSON：}\\
\textit{(Strict JSON output:)}\\
\texttt{\{ "l1": "slogan|substantive|irrelevant", "l2": "",
"confidence\_substantive": 0.0, "stance\_scores":
\{...five keys...\}, "slogan\_density": 0.0 \}}\\[3pt]
\zh{confidence\_substantive：是 substantive 的概率（0-1）}\\
\zh{stance\_scores[dim]：假设这段是 substantive 时该维度的 stance 强度（0-1）。维度词出现 ≠ 高分，要看具体行动/数字/取舍}\\
\zh{slogan\_density：套话密度（Xi 引文+四字格+抽象口号 占段比例，0-1）}\\
\textit{(Field semantics: confidence\_substantive is P(segment is
substantive); stance\_scores[dim] is stance strength on the dimension
\emph{conditional on substantive}; the presence of dimension keywords
does not entail a high score, what matters is concrete actions, numbers,
and tradeoffs; slogan\_density is the character fraction occupied by
political set phrases, in [0,1].)}\\[3pt]
\zh{段落：} \texttt{\{segment\_text\}}\\
\textit{(Segment: [substituted at run time])}
\end{quote}

\section{Full $\lambda$ grid}
\label{app:grid}

\begin{table}[h]
\centering
\small
\begin{tabular}{r r r r r}
\toprule
$\lambda_{\text{llm}}$ & $\lambda_{\text{ng}}$ & $\bar D_{\text{lc}}$ & $\bar D_{\text{sl}}$ & $d$ \\
\midrule
0.0 & 0.0 & 0.178 & 0.045 & 0.86 \\
0.0 & 1.0 & 0.178 & 0.045 & 0.86 \\
0.0 & 2.0 & 0.178 & 0.045 & 0.86 \\
0.0 & 3.0 & 0.178 & 0.045 & 0.86 \\
0.5 & 0.0 & 0.179 & 0.047 & 0.85 \\
0.5 & 2.0 & 0.178 & 0.047 & 0.85 \\
1.0 & 2.0 & 0.179 & 0.049 & 0.83 \\
1.0 & 3.0 & 0.179 & 0.049 & 0.83 \\
1.5 & 2.0 & 0.179 & 0.050 & 0.83 \\
2.0 & 3.0 & 0.179 & 0.051 & 0.83 \\
\bottomrule
\end{tabular}
\vspace{-3mm}\caption{Selected entries from the $\lambda$ grid search
(§\ref{sec:results}).  The landscape is flat: across all 20 grid
points, Cohen $d$ remains within $[0.83, 0.86]$.}
\label{tab:grid}
\end{table}

\section{Slogan lexicon (full)}
\label{app:lexicon}

The full 53-entry auto-mined slogan lexicon, along with the six
high-frequency entries shown in Table~\ref{tab:slogan}, is released
with the corpus.  Entries are jieba 5-grams that appear in at least
15\% of the 80 documents.  Examples not shown in the main text include
the four-character formulations \emph{``two unwaverings''} (the dual
commitment to public-sector and non-public-sector economic
development), \emph{``new development concept''}, and recurring
references to ``the 18th Party Congress'' and onwards.

\section{Paraphrase robustness table}
\label{app:para}

\begin{table}[h]
\centering
\small
\begin{tabular}{lrrr}
\toprule
Method & mean$_{\text{orig}}$ & mean$_{\text{rew}}$ & retention \\
\midrule
\textsc{Dict}       & 0.020 & \textbf{0.027} & \textbf{1.55} \\
\textsc{Llm} (raw)  & 0.749 & 0.623 & 0.75 \\
\textsc{Calibrated} & 0.590 & 0.453 & \textbf{0.69} \\
\bottomrule
\end{tabular}
\vspace{-3mm}\caption{Paraphrase robustness on 50 substantive-to-slogan rewrites
(§\ref{sec:para}).  mean$_{\text{orig}}$ and mean$_{\text{rew}}$ are
mean per-segment maximum-dimension scores; retention is the mean
per-pair ratio of rewrite to original.  Stance-aware methods should
retain less than 1; surface methods retain near 1 (or above, if the
rewrite is more keyword-dense than the original).}
\label{tab:para}
\end{table}

\section{Cross-LLM agreement details}
\label{app:crossllm}

We report two cross-LLM checks against Qwen3.5:9b (the model used in
the main run), each on an 85-segment stratified subsample (see
§\ref{sec:crossllm}).

\paragraph{Within-family (Qwen3.5:27b).}  L1 agreement 86\%;
$\kappa_{L1} = 0.746$; $\kappa_{L2} = 0.510$ on the 45
mutually-\textsc{substantive} segments.  Per-dimension Pearson $r$
on raw stance scores: \textit{Innovation} $0.900$,
\textit{Competition--Cooperation} $0.816$,
\textit{Organisation--Market} $0.921$,
\textit{Social Responsibility} $0.929$, \textit{National Mission}
$0.844$; mean $0.882$.  Slogan density correlates at $r = 0.729$ and
substantive confidence at $r = 0.890$.

\paragraph{Cross-family (DeepSeek-r1:8b).}  L1 agreement 83\%;
$\kappa_{L1} = 0.695$; $\kappa_{L2} = 0.297$ on the 50
mutually-\textsc{substantive} segments.  Per-dimension Pearson $r$:
\textit{Innovation} $0.846$, \textit{Competition--Cooperation}
$0.727$, \textit{Organisation--Market} $0.714$,
\textit{Social Responsibility} $0.779$, \textit{National Mission}
$0.752$; mean $0.764$.  Slogan density correlates at $r = 0.719$ and
substantive confidence at $r = 0.844$.  Cross-family agreement is
weaker than within-family but still substantial on L1 and the
confidence channel, suggesting that the slogan-vs-substance
distinction is recovered consistently across architecture families.

\section{Sensitivity analyses (full)}
\label{app:sensitivity}

\paragraph{Leave-one-SL-out.}
Dropping each of the 5 same-leader pairs and recomputing
($n_{\text{lc}}=24$, $n_{\text{sl}}=4$): \textsc{Llm} Cohen $d \in
[0.98, 1.25]$, perm $p_1 \in [0.024, 0.071]$; \textsc{Calibrated}
$d \in [0.76, 0.94]$, $p_1 \in [0.029, 0.100]$; \textsc{Dict}
$d \in [0.73, 0.98]$, $p_1 \in [0.072, 0.142]$; \textsc{Lda}
$d \in [0.08, 0.93]$, $p_1 \in [0.048, 0.457]$; \textsc{Bge}
$d \in [0.55, 0.95]$, $p_1 \in [0.020, 0.262]$.

\paragraph{Placebo (random SL/LC reassignment).}
2000 random partitions of the 29 doc pairs into a size-5 ``SL''
group and a size-24 ``LC'' group; report the fraction of placebo
trials with Cohen $d \geq$ observed.  \textsc{Llm} $3.9\%$,
\textsc{Calibrated} $3.5\%$, \textsc{Dict} $7.4\%$, \textsc{Bge}
$14.4\%$, \textsc{Lda} $35.7\%$.  Only the LLM methods reject the
null at $\alpha=0.05$.

\paragraph{Style-residualised paired contrast.}
Per-document dimension scores are residualised against five
stylometric features (sentence-length mean and SD, numeric density,
long-run density as a named-entity proxy, character type-token
ratio).  Paired contrast on residuals: \textsc{Llm} $d=0.43$, perm
$p_1=0.22$; \textsc{Calibrated} $d=0.39$, $p_1=0.23$; \textsc{Bge}
$d=0.71$, $p_1=0.12$; \textsc{Dict} $d=0.51$, $p_1=0.16$;
\textsc{Lda} $d=0.35$, $p_1=0.26$.  Absolute $\Delta$ remains
positive for \textsc{Llm} ($0.22$) and \textsc{Calibrated} ($0.19$);
the rank ordering of methods is preserved but no method is
significant at $\alpha=0.05$ after this aggressive residualisation.

\section{Case-study paragraph pair}
\label{app:case}

\begin{table*}[h]
\centering
\footnotesize
\begin{tabular}{p{0.45\textwidth} p{0.50\textwidth}}
\toprule
\textbf{Slogan paragraph (high keyword density)} & \textbf{Substantive paragraph (concrete actions)} \\
\midrule
\textit{``We will adhere to high-quality development \dots strengthen
competitiveness, innovation capability, control, influence, and
risk-resistance of state-owned capital.''} (translated; SASAC
chairperson) &
\textit{``We took 4\,Mt of the cement market in 2018; before
restructuring, the segment ran at a loss. After the Wang-Zhuang
roundtable in 2007 we consolidated 150 cement producers into Southern
Cement; 2018 after-tax profit was 10\,bn yuan.''} (translated; CNBM
chairperson) \\
\midrule
\textsc{Dict} hits: 5 keywords; predicted dominant dimension:
\emph{Innovation} (score 0.18) &
\textsc{Dict} hits: 1 keyword; predicted dominant dimension:
\emph{Organisation--Market} (score 0.05) \\
\midrule
\textsc{Llm} L1 = \textsc{slogan}; substantive confidence 0.08;
\textsc{Calibrated} \emph{Innovation} score: 0.02 &
\textsc{Llm} L1 = \textsc{substantive}; substantive confidence 0.95;
\textsc{Calibrated} \emph{Organisation--Market} score: 0.71 \\
\bottomrule
\end{tabular}
\vspace{-3mm}\caption{The qualitative case-study paragraph pair referenced in
§\ref{sec:case}.  The dictionary baseline scores the slogan paragraph
as more strongly indicative of \emph{Innovation} than the substantive
paragraph is of \emph{Organisation--Market}.  The \textsc{Llm}'s
substantive-confidence multiplier collapses the slogan paragraph's
stance vector while preserving the substantive paragraph's.}
\label{tab:case}
\end{table*}

\end{document}